\documentclass{article}
\usepackage{ijcai24}

\usepackage{times}
\usepackage{soul}
\usepackage{url}
\usepackage[hidelinks]{hyperref}
\usepackage[utf8]{inputenc}
\usepackage[small]{caption}
\usepackage{graphicx}
\usepackage{amsmath}
\usepackage{amsthm}
\usepackage{booktabs}
\usepackage{subfigure}
\usepackage{multirow}
\usepackage[switch]{lineno}

\usepackage[ruled,vlined]{algorithm2e}

\usepackage{xcolor}

\title{Rethinking the Soft Conflict Pseudo Boolean Constraint \\on MaxSAT 
Local Search Solvers
}

\newcommand{\name}[0]{SPB-MaxSAT\xspace}

\author{
    Jiongzhi Zheng$^1$\equalcontrib \and  
    Zhuo Chen$^1$\equalcontrib \and    
    Chu-Min Li$^2$\and 
    Kun He$^1$\thanks{Corresponding author.}
    \affiliations
    $^1$School of Computer Science and Technology, Huazhong University of Science and Technology, China\\
    $^2$MIS, University of Picardie Jules Verne, France\\
    \emails
    brooklet60@hust.edu.cn
}

\begin{document}

\maketitle

\begin{abstract}
MaxSAT is an optimization version of the famous NP-complete Satisfiability problem (SAT). Algorithms for MaxSAT mainly include complete solvers and local search incomplete solvers. In many complete solvers, once a better solution is found, a Soft conflict Pseudo Boolean (SPB) constraint will be generated to enforce the algorithm to find better solutions. In many local search algorithms, clause weighting is a key technique for effectively guiding the search directions. In this paper, we propose to transfer the SPB constraint into the clause weighting system of the local search method, leading the algorithm to better solutions. We further propose an adaptive clause weighting strategy that breaks the tradition of using constant values to adjust clause weights. Based on the above methods, we propose a new local search algorithm called \name that provides new perspectives for clause weighting on MaxSAT local search solvers. Extensive experiments demonstrate the excellent performance of the proposed methods.
\end{abstract}

\section{Introduction}
\label{sec-Intro}
The Maximum Satisfiability problem (MaxSAT) is an optimization version of the famous Satisfiability problem (SAT). Given a propositional formula in Conjunctive Normal Form (CNF), MaxSAT aims to find an assignment that satisfies as many clauses as possible. More generally, the Partial MaxSAT (PMS) divides the clauses into hard and soft, aiming to satisfy as many soft clauses as possible while satisfying all the hard clauses. The Weighted PMS (WPMS) associates each soft clause with a positive weight, aiming to maximize the total weight of satisfied soft clauses while satisfying all the hard clauses. The nature of hard and soft clauses well matches the constraints and optimization objectives in optimization scenarios. Therefore, MaxSAT has a wide range of industrial and academic applications, such as group testing~\cite{GroupTesting}, timetabling~\cite{Timetabling}, planning~\cite{Planning}, clique problems~\cite{Clique}, set covering~\cite{SetCover}, etc.

Algorithms for MaxSAT can be divided into complete and incomplete solvers based on their capacity to provide optimality guarantees. Complete algorithms include branch and bound~\cite{BnB1,MaxCDCL} algorithms and a series of methods that solve MaxSAT by iteratively calling a SAT solver, so-called SAT-based algorithms~\cite{SAT-based1,SAT-based2,SAT-based3}. Due to the power of complete solving techniques in handling CNF formulas, such as the unit propagation~\cite{DPLL,UP-ChuMinLi} and clause learning~\cite{CDCL,Vivification}, complete algorithms exhibit excellent performance for solving MaxSAT.

Another important technique in MaxSAT complete-solving is the Soft conflict Pseudo Boolean (SPB) constraints~\cite{Loandra,MaxCDCL}. A Pseudo Boolean (PB) constraint~\cite{PBO1,PBO2} is a particular type of linear constraint in the form of $\sum_{i=1}^n q_ix_i~\text{\small{\#}}~K$, where $\text{\small{\#}}\in \{<,\leq, =,\geq, > \}$ and can be normalized to $<$, $q_i$ and $K$ are integer constants, and $x_i$ are 0/1 variables, $i\in \{1,...,n\}$. In many complete MaxSAT solvers, once a better solution is found, a PB (or Cardinality for unweighted problem) constraint~\cite{PBconstraints} is added naturally to prohibit the procedure from finding solutions that are no better than the current best. Falsifying such a constraint leads to a Soft conflict~\cite{MaxCDCL}. The SPB constraints can help complete solvers prune branches and reduce the search space.

On the other hand, incomplete algorithms mainly focus on the local search approach~\cite{LS1,LS2,LS3}. With the development of many local search techniques, including the decimation-based initialization methods~\cite{Decimation}, the local optima escaping methods~\cite{BandMaxSAT,FPS}, and especially the clause weighting methods~\cite{SATLike,SATLike3.0,NuWLS}, the state-of-the-art local search algorithms exhibit competitive performance with complete algorithms and even dominating the rankings of incomplete tracks of recent MaxSAT Evaluations.


In this paper, we focus on the local search approach for MaxSAT and propose two new methods regarding the clause weighting scheme, i.e., integrating SPB constraints into the clause weighting system and a novel Adaptive Clause Weighting scheme.



We first introduce the SPB constraints into the local search MaxSAT algorithm. Similar constraints have also been applied to local search for Pure MaxSAT~\cite{LinearLS}, a special case of MaxSAT 
whose hard clauses have only positive literals and soft clauses are all unit clauses with negative literals (and vice versa), and it is used as a hard constraint to strictly restrict the search space, which may limit the search ability of the local search.

To better and more generally make use of the SPB constraints, we propose to integrate them into the clause weighting system for universal MaxSAT problems, including PMS and WPMS. Specifically, we associate each hard clause with a dynamic weight and add a 
SPB constraint to the MaxSAT formula with a dynamic weight. Once the algorithm falls into a local optimum, the dynamic weights of the hard clauses and the SPB constraint falsified by the current local optimal solution will be increased. Moreover, once a better solution is found by local search, the SPB constraint will be updated accordingly. In this way, the SPB constraint does not restrict the search space but guides the search directions and leads the algorithm to find better solutions.

For the second method, we found that the existing clause weighting methods usually update the dynamic weight of each clause using a constant value, which may make the increased rate of dynamic weights converge inversely proportional to the number of weight increments and approaches zero. The convergence of the increased rate may lead to the distribution of dynamic weights tending to stabilize, diminish the effect of the clause weighting method, and make the algorithm hard to escape from local optima. 
There are some techniques for avoiding the convergence of the increased rate caused by the divergent increased dynamic weights, including setting upper limits, periodic decay, and probabilistic smoothing~\cite{CCAnr,SATLike}. However, setting upper limits may restrict the 
search flexibility, 
the decay method cannot handle the convergence during each decay, and the weight smoothing method should set lower limits, which may also restrict the 
search flexibility.

To handle these issues, we propose a simple clause weighting method that allows the dynamic weights to update adaptively. Specifically, we increase the dynamic weights proportionally according to the current weights. Such a strategy provides a lower bound for the increased rate of dynamic weights and makes the clause weighting method always guide the search directions and help escape from local optima efficiently. As a result, our method is simple yet effective, providing a new perspective on the clause weighting methods.

In the end, by integrating the above two methods, we propose a new local search MaxSAT algorithm called \name. \name always maintains an SPB constraint and associates it with a dynamic weight, which will be increased proportionally once a local optimal solution falsifies the constraint. We further combine \name with a typical SAT-based solver, TT-Open-WBO-Inc~\cite{SAT-based3}, as many incomplete MaxSAT solvers do~\cite{SATLike-c,NuWLS-c}, and the resulting incomplete solver is denoted as \name-c. Extensive experiments show that \name significantly outperforms the state-of-the-art local search MaxSAT algorithms, NuWLS~\cite{NuWLS} and BandMaxSAT~\cite{BandMaxSAT}; \name-c also exhibits higher performance and robustness than NuWLS-c-2023~\cite{NuWLS-c-2023}, winner of all the four incomplete tracks of MaxSAT Evaluation 2023.


\section{Related Work}
This section first reviews the development of the clause weighting method in local search MaxSAT algorithms and then introduces related works that use complete solving techniques of SAT and MaxSAT for the local search. We will also briefly describe the differences and improvements of our method over the existing studies.

\subsection{Clause Weighting in MaxSAT Local Search}
Clause weighting is a 
typical technique in local search SAT and MaxSAT solvers~\cite{Godot,LS3}. It associates the clauses with dynamic weights and uses them to guide the search directions. The main idea is that when the algorithm falls into a local optimum, i.e., flipping any variable cannot improve the current solution, the dynamic weights of falsified clauses will be increased to help the algorithm escape from local optima. With the appearance of PMS and WPMS, many advanced clause weighting methods have been proposed to consider the properties of hard and soft clauses. For instance, Dist~\cite{Dist} and CCEHC~\cite{CCEHC} only use clause weighting methods upon hard clauses and prioritize satisfying hard clauses during the search. SATLike~\cite{SATLike} and its extension of SATLike3.0~\cite{SATLike3.0} first propose to associate both hard and soft clauses with dynamic weights, which is also inherited by BandMaxSAT~\cite{BandMaxSAT} and FPS~\cite{FPS} algorithms. Recently, NuWLS~\cite{NuWLS} improves the clause weighting method by assigning appropriate initial dynamic weights to the clauses, and NuWLS-c-2023~\cite{NuWLS-c-2023} further proposes to uniformly update the dynamic weights of all soft clauses, helping it win all the four incomplete tracks of MaxSAT Evaluation 2023.

The clause weighting method has made significant achievements in the field of MaxSAT solving. In this work, we investigate the limitation of constant increments of existing clause weighting methods used in the above algorithms and thereby propose an effective adaptive clause weighting strategy to address the limitation.

\subsection{Complete Solving Methods in Local Search}
Since complete algorithms occupy the mainstream of (Max)SAT solvers, researchers usually use local search to improve complete solvers~\cite{CDCL-LS,NuWLS-c-2023} but rarely try to use complete solving methods to improve local search incomplete solvers. There are some studies that use unit propagation~\cite{DPLL,UP-ChuMinLi} techniques to generate the initial assignments for local search (Max)SAT algorithms, obtaining significant improvements. The linear search constraint has been used in the local search process by strictly restricting the search space for a special case of MaxSAT~\cite{LinearLS}, 
showing excellent performance in the special MaxSAT problems that are very easy to find feasible solutions. However, strictly restricting the search space might not be appropriate for general MaxSAT problems.

This paper also investigates the utilization of complete solving methods, i.e., SPB constraint, in MaxSAT local search solvers and proposes to integrate SPB into the clause weighting system rather than directly restrict the search space. The clause weighting scheme with the SPB constraint can guide the algorithm to find better solutions, 
as demonstrated by our follow-up experiments on various MaxSAT problems.

\section{Preliminaries}
Given a set of Boolean variables $\{x_1, \cdots, x_n\}$, a literal is either a variable itself $x_i$ or its negation $\lnot x_i$, a clause is a disjunction of literals, e.g., $c = l_1 \lor \cdots \lor l_p$, and a Conjunctive Normal Form (CNF) formula $\mathcal{F}$ is a conjunction of clauses, e.g., $\mathcal{F} = c_1 \land \cdots \land c_q$. A complete assignment $A$ represents a mapping that maps each variable to a value of 1 (true) or 0 (false). A literal $x_i$ (resp. $\lnot x_i$) is satisfied if the current assignment maps $x_i$ to 1 (resp. 0). A clause is satisfied by the current assignment if there is at least one satisfied literal in the clause and is falsified otherwise.

Given a CNF formula $\mathcal{F}$, the MaxSAT problem aims to find an assignment (i.e., solution) that satisfies as many clauses in $\mathcal{F}$ as possible. Given a CNF formula $\mathcal{F}$ whose clauses are divided into hard and soft, we denote $Hard(\mathcal{F})$ and $Soft(\mathcal{F})$ as the set of hard and soft clauses in $\mathcal{F}$, respectively. The Partial MaxSAT (PMS) problem is a variant of MaxSAT that aims to find an assignment satisfying all the hard clauses while maximizing the number of satisfied soft clauses in $\mathcal{F}$. The Weighted Partial MaxSAT (WPMS) problem is a generalization of PMS where each soft clause is associated with a positive weight, aiming to find an assignment that satisfies all the hard clauses while maximizing the total weight of satisfied soft clauses in $\mathcal{F}$.

For convenience, we regard PMS as a special case of WPMS by assigning each soft clause with unit weight. Given a MaxSAT instance $\mathcal{F}$, we denote $w_{s}(c)$ as the original weight of soft clause $c \in Soft(\mathcal{F})$ and denote $w_{h}(c)$ as the dynamic weight maintained of each hard clause $c \in Hard(\mathcal{F})$ by the clause weighting method.

Given a MaxSAT instance $\mathcal{F}$, a complete assignment $A$ is feasible if it satisfies all the hard clauses in $\mathcal{F}$, and we denote its objective function 
$obj(A)$ as the total weight of soft clauses falsified by $A$ and set its cost function $cost(A)$ to $obj(A)$ (resp. $+\infty$) if $A$ is feasible (resp. infeasible). 
Moreover, in the local search algorithms for MaxSAT, the widely-used flipping operator for a variable is an operator that changes its Boolean value.




\section{The Proposed \name}
This section introduces our proposed local search algorithm, termed the Soft conflict Pseudo Boolean based MaxSAT solver (\name). \name maintains an SPB constraint with dynamic weight during the local search process, integrating SPB into the clause weighting system and using an adaptive clause weighting technique to update its dynamic weight. The clause weighting system is then used to help the algorithm escape from local optima and find better solutions.

In the following, we first introduce the clause weighting system in \name with the SPB constraint and the adaptive clause weighting method and then present the main framework of \name.


\subsection{Clause Weighting System}
This subsection presents our proposed SPB constraint, scoring function, and adaptive SPB weighting method together with our intuition. 

\subsubsection{SPB Constraint}
Suppose $A^*$ is the best solution found so far during the local search. Then, finding solutions no better than $A^*$ is meaningless, and an SPB constraint, denoted as $SPB$, is added naturally as follows.

\begin{equation}
SPB: \sum_{c \in Soft({\mathcal{F}})}{w_{s}(c) \cdot unsat(c) < cost(A^*)},
\label{eq:PB}
\end{equation}
where $unsat(c)$ equals 1 (resp. 0) if clause $c$ is falsified (resp. satisfied) by the current solution $A$ and can be regarded as a Boolean variable. Actually, $SPB$ can be simply represented by $obj(A) < cost(A^*)$, and it will be updated accordingly with the updating of $A^*$.

In \name, we associate $SPB$ with a dynamic weight $w(SPB)$, using $w(SPB)$ and the dynamic weights of hard clauses to calculate the scoring function of the variables and guide the search directions.

\subsubsection{Scoring Function}
In the local search MaxSAT algorithms, the scoring function $score(v)$ of a variable $v$ is widely used to evaluate the benefit of flipping $v$ in the current solution. In our \name, $score(v)$ is calculated by combining the influences of 
such a flipping on the hard clauses and the $SPB$ constraint.

Given a MaxSAT instance $\mathcal{F}$, the current solution $A$, the dynamic weight $w_h(c)$ of each hard clause $c$, and the dynamic weight $w(SPB)$ of $SPB$, we denote $hscore(v)$ as the decrement of the total dynamic weight of falsified hard clauses caused by flipping $v$ in $A$. Suppose $A'$ is the solution obtained by flipping variable $v$ in $A$. We define another scoring function, $SPBscore(v)$, to evaluate the influence of flipping $v$ in $A$ to the $SPB$ constraint as follows.
\begin{equation}
SPBscore(v) = w(SPB) \cdot (obj(A) - obj(A')).
\label{eq:SPBscore}
\end{equation}

Finally, the scoring function $score(v)$ is defined as follows:
\begin{equation}
score(v) = hscore(v) + SPBscore(v).
\label{eq:score}
\end{equation}

The dynamic weights $w_h(c)$ and $w(SPB)$ actually play the role of a pivot to control the importance of the hard clauses and soft clauses in the MaxSAT solving process. The $SPBscore(v)$ actually regards all soft clauses as a whole and evaluates the influence on the objective function $obj(A)$ caused by flipping $v$. In our method, the clause weighting technique is used to adjust $w_h(c)$ and $w(SPB)$, so as to adjust the precedence of satisfying hard clauses and increasing the objective function. As a result, the adaptively adjusted $SPB$ Constraint and $w(SPB)$ dynamic weight can lead the local search algorithm to better solutions.


\subsubsection{Adaptive SPB Weighting}
The initial values of the dynamic weight $w_h(c)$ of each hard clause $c$ and the dynamic weight $w(SPB)$ of $SPB$ are both set to 1. We propose an SPB-Weighting function shown in Algorithm~\ref{alg:SPB-Weighting} to update the dynamic weights once the algorithm falls into a local optimum, i.e., there is no variable having a positive $score$. 

\begin{algorithm}[t]
\fontsize{10.1pt}{15}
\caption{SPB-Weighting}
\label{alg:SPB-Weighting}
\LinesNumbered 
\KwIn{MaxSAT instance $\mathcal{F}$, current solution $A$, hard clause dynamic weight increment $h_{inc}$, increment proportion $\delta$} 
\For{each clause $c \in Hard(\mathcal{F})$ falsified by $A$}{
$w_h(c) \leftarrow w_h(c) + h_{inc}$\;
}
\If{$SPB$ is falsified by $A$}{
$w(SPB) \leftarrow \delta \cdot (w(SPB) + 1)$\;
}
\end{algorithm}


\begin{figure}[!t]
\centering
\vspace{-1em}
\includegraphics[width=1.0\columnwidth]{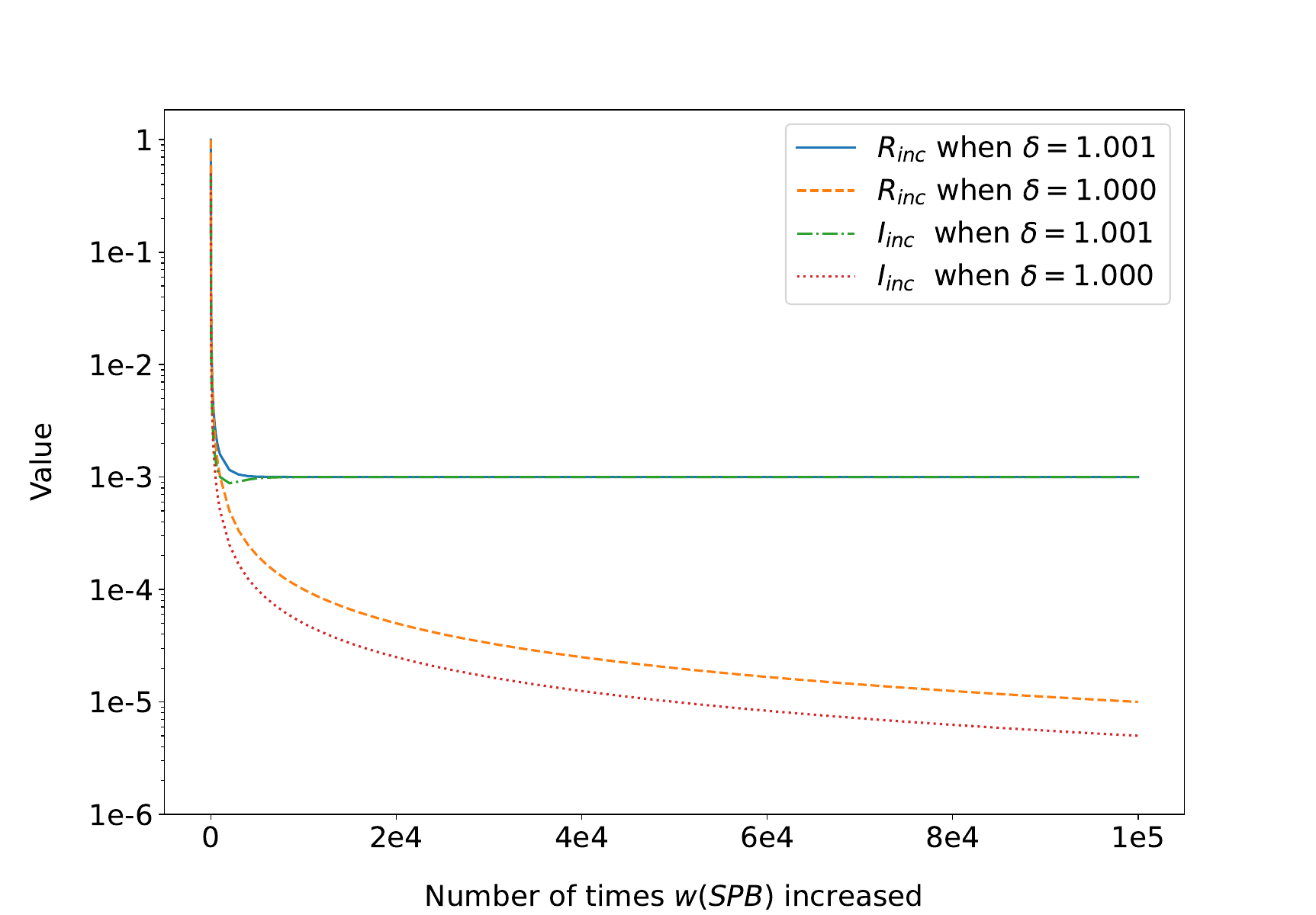}
\caption{How $R_{inc}$ and $I_{inc}$ change with the increment of the number of times $w(SPB)$ increases.}
\label{fig:analysis}
\end{figure}

The SPB-Weighting function is very simple and easy to understand. This function just increases the dynamic weights of falsified elements in the current local optimal solution, including hard clauses and the $SPB$ constraint. The dynamic weights of falsified hard clauses are increased by a constant parameter $h_{inc}$ as many local search MaxSAT algorithms do~\cite{SATLike,BandMaxSAT,NuWLS}, and the dynamic weight of $SPB$ is updated adaptively by first increasing 1 and then multiplying by $\delta > 1$. Actually, the original constant clause weighting method can be regarded as a special case of our adaptive clause weighting method with $\delta = 1$. Moreover, to avoid $w(SPB)$ from becoming too large, we also use a decay method to reduce all the dynamic weights simultaneously when they reach a threshold. 

\begin{algorithm}[t]
\fontsize{10.1pt}{15}
\caption{\name}
\label{alg:main}
\LinesNumbered 
\KwIn{MaxSAT instance $\mathcal{F}$, cut-off time \textit{cutoff}, BMS parameter $k$, hard clause dynamic weight increment $h_{inc}$, increase proportion $\delta$}
\KwOut{A feasible solution $A$ of $\mathcal{F}$, or \textit{no feasible solution found}}
$A \leftarrow$ an initial complete assignment\;
$A^* \leftarrow A$\;
\For{each $c \in Hard(\mathcal{F})$}{$w_h(c) \leftarrow 1$\;}
$w(SPB) \leftarrow 1$\;
\While{running time $<$ cutoff}{
\eIf{$D \leftarrow \{x|score(x)>0\} \neq \emptyset$}{
$v \leftarrow$ a variable in $D$ picked by BMS($k$)\;
}
{Update dynamic weights by SPB-Weighting\;
\eIf{$\exists$ falsified hard clauses}{
$c \leftarrow$ a random falsified hard clause\;
}{
$c \leftarrow$ a random falsified soft clause\;
}
$v \leftarrow$ the variable with the highest $score$ in $c$\;
}
$A \leftarrow A$ with $v$ flipped\;
\If{$cost(A) < cost(A^*)$}{
$A^* \leftarrow A$\;
update $SPB$ accordingly\;
}
}
\lIf{$A^*$ is feasible}{\textbf{return} $A^*$}
\lElse{\textbf{return} \textit{no feasible assignment found}}
\end{algorithm}

\subsubsection{Discussion on Adaptive Clause Weighting}
The adaptive clause weighting method in function SPB-Weighting actually highlights the soft clauses and better solutions in the later stages of the search. At the beginning of the search, when there is no feasible solution found, $cost(A^*) = +\infty$ holds, $SPB$ is always satisfied, and $w(SPB)$ always equals 1. The algorithm mainly pays attention to finding feasible solutions. After finding a feasible solution, $w(SPB)$ will be increased once $SPB$ is falsified to pay more attention to soft clauses and push the algorithm to find better solutions. However, with the accumulation of the dynamic weights, increasing $w(SPB)$ by 1 might only have a minor influence on the importance of soft clauses, and the algorithm might need more steps to 
pay appropriate attention to the soft clauses.

Suppose the average dynamic weights of hard clauses, denoted as $w_h(\overline{c})$, increases by 1 once $w(SPB)$ increases. For each increment of $w(SPB)$, i.e., $w'(SPB) = \delta \cdot (w(SPB) + 1)$, we define the increasing rate as $R_{inc} = (w'(SPB) - w(SPB)) / w(SPB)$ and the increment on the importance of soft clauses as $I_{inc} = (w'(SPB) - w(SPB))/(w(SPB)+w_h(\overline{c}))$. The smaller the $R_{inc}$ and $I_{inc}$, the smaller the effect of the increment of $w(SPB)$ for guiding the search. Figure \ref{fig:analysis} shows how these two metrics change with the increment of the number of times $w(SPB)$ increased when $\delta = 1.001$ (i.e., adaptive weighting) and 1 (i.e., constant weighting). We can observe that both the two metrics converge to $\delta - 1$, i.e., zero when $\delta = 1$. Therefore, increasing $w(SPB)$ proportionally with $\delta > 1$ can avoid the diminishing of the clause weighting method for guiding the search with the accumulation of the dynamic weights.

In addition, for the hard clauses, we increase their dynamic weights conservatively to ensure that we can find high-quality feasible solutions. 
The experimental results also show that increasing $w(SPB)$ proportionally is better than increasing both $w_h(c)$ and $w(SPB)$ proportionally or linearly with a constant value, demonstrating that our design is reasonable and effective.

\subsection{The Main Framework}
The main procedure of \name is shown in Algorithm~\ref{alg:main}. The algorithm first generates an initial complete assignment $A$ by the decimation method~\cite{Decimation} used in SATLike3.0~\cite{SATLike3.0} and NuWLS~\cite{NuWLS}, initializes the best solution found $A^*$, and initializes the dynamic weights (lines 1-5). Then, the algorithm repeats selecting a variable and flipping it until the cut-off time is reached (lines 6-19). If there are variables with positive $score$, the algorithm also uses the Best-from-Multiple-Selections (BMS) strategy~\cite{BMS} used in various local search MaxSAT algorithms to select a variable with positive $score$ (lines 7-8). BMS chooses $k$ random variables with positive $score$ (with replacement) and returns the one with the highest $score$.

Upon falling into a local optimum, \name first updates the dynamic weights by the SPB-Weighting function (line 10). Then, if the current local optimal solution is infeasible (resp. feasible), a random falsified hard (resp. soft) clause $c$ is selected, and the variable to be flipped is the one with the highest $score$ in $c$ (lines 11-15). Once a better feasible solution is found, $A^*$ will be updated, and the $SPB$ constraint will also be updated accordingly.

\section{Experiments}
In this section, we first compare the proposed SPB-MaxSAT\footnote{The source code of \name is available at https://github.com/[MASKED-FOR-REVIEW].} algorithm with the state-of-the-art local search algorithms, NuWLS~\cite{NuWLS} and BandMaxSAT~\cite{BandMaxSAT}. We then combine \name with a SAT-based solver, TT-Open-WBO-Inc~\cite{SAT-based3}, as many incomplete MaxSAT solvers do~\cite{SATLike-c,NuWLS-c}, and compare the resulting incomplete solver \name-c with the state-of-the-art incomplete solver NuWLS-c-2023, which won all the four incomplete tracks of MaxSAT Evaluation (MSE) 2023. NuWLS-c-2023 is an improvement of NuWLS-c~\cite{NuWLS-c}, the winner of all the four incomplete tracks of MSE2022. Since they obtained such good results on recent MSEs, we only selected NuWLS-c-2023 as the baseline of incomplete solver and 
ignored other effective solvers, such as DT-HyWalk~\cite{DT-HyWalk}, Loandra~\cite{Loandra}, and TT-Open-WBO-Inc~\cite{SAT-based3}. Finally, we perform ablation studies to analyze the effectiveness of our adaptive clause weighting method by comparing \name with its variant algorithms.

\begin{table*}[t]
\footnotesize
\centering
\begin{tabular}{lrrrrrrrrr|rrrrrrrr} \toprule
\multirow{2}{*}{Benchmark} &  \multirow{2}{*}{\hspace{-0.55em}\#inst.} &  \multicolumn{3}{c}{\name (60s)}     & \multicolumn{1}{c}{\hspace{-0.55em}} &  \multicolumn{3}{c}{NuWLS (60s)} &  \multicolumn{1}{c}{\hspace{-0.55em}} &  \multicolumn{1}{|c}{\hspace{-0.55em}} &  \multicolumn{3}{c}{\name (300s)}     &  \multicolumn{1}{c}{\hspace{-0.55em}} &  \multicolumn{3}{c}{NuWLS (300s)} \\ \cline{3-5} \cline{7-9} \cline{12-14} \cline{16-18} 
                           & \hspace{-0.55em} \hspace{-0.55em}                          & \hspace{-0.55em} \hspace{-0.55em} \#win       & \hspace{-0.55em} \hspace{-0.55em} time  & \hspace{-0.55em} \hspace{-0.55em} \#score        & \hspace{-0.55em} \hspace{-0.55em}                      & \hspace{-0.55em} \hspace{-0.55em} \#win   & \hspace{-0.55em} \hspace{-0.55em} time    & \hspace{-0.55em} \hspace{-0.55em} \#score   & \hspace{-0.55em} \hspace{-0.55em}                      & \hspace{-0.55em} \hspace{-0.55em}                      & \hspace{-0.55em} \hspace{-0.55em} \#win       & \hspace{-0.55em} \hspace{-0.55em} time   & \hspace{-0.55em} \hspace{-0.55em} \#score        & \hspace{-0.55em} \hspace{-0.55em}                      & \hspace{-0.55em} \hspace{-0.55em} \#win    & \hspace{-0.55em} \hspace{-0.55em} time    & \hspace{-0.55em} \hspace{-0.55em} \#score   \\ \hline
PMS2018                    & \hspace{-0.55em} 153                      & \hspace{-0.55em} \textbf{115} & \hspace{-0.55em} 15.58 & \hspace{-0.55em} \textbf{0.7562} & \hspace{-0.55em} \textbf{}            & \hspace{-0.55em} 61       & \hspace{-0.55em} 17.34   & \hspace{-0.55em} 0.7224     & \hspace{-0.55em}                      & \hspace{-0.55em}                      & \hspace{-0.55em} \textbf{118} & \hspace{-0.55em} 63.14  & \hspace{-0.55em} \textbf{0.7819} & \hspace{-0.55em} \textbf{}            & \hspace{-0.55em} 70        & \hspace{-0.55em} 60.09   & \hspace{-0.55em} 0.7583     \\
PMS2019                    & \hspace{-0.55em} 299                      & \hspace{-0.55em} \textbf{230} & \hspace{-0.55em} 13.54 & \hspace{-0.55em} \textbf{0.7384} & \hspace{-0.55em} \textbf{}            & \hspace{-0.55em} 142      & \hspace{-0.55em} 12.22   & \hspace{-0.55em} 0.7205     & \hspace{-0.55em}                      & \hspace{-0.55em}                      & \hspace{-0.55em} \textbf{235} & \hspace{-0.55em} 56.75  & \hspace{-0.55em} \textbf{0.7701} & \hspace{-0.55em} \textbf{}            & \hspace{-0.55em} 150       & \hspace{-0.55em} 42.72   & \hspace{-0.55em} 0.7491     \\
PMS2020                    & \hspace{-0.55em} 262                      & \hspace{-0.55em} \textbf{198} & \hspace{-0.55em} 14.22 & \hspace{-0.55em} \textbf{0.7378} & \hspace{-0.55em} \textbf{}            & \hspace{-0.55em} 113      & \hspace{-0.55em} 13.78   & \hspace{-0.55em} 0.7156     & \hspace{-0.55em}                      & \hspace{-0.55em}                      & \hspace{-0.55em} \textbf{199} & \hspace{-0.55em} 60.41  & \hspace{-0.55em} \textbf{0.7586} & \hspace{-0.55em} \textbf{}            & \hspace{-0.55em} 127       & \hspace{-0.55em} 53.71   & \hspace{-0.55em} 0.7454     \\
PMS2021                    & \hspace{-0.55em} 155                      & \hspace{-0.55em} \textbf{119} & \hspace{-0.55em} 16.61 & \hspace{-0.55em} \textbf{0.6565} & \hspace{-0.55em} \textbf{}            & \hspace{-0.55em} 70       & \hspace{-0.55em} 6.82    & \hspace{-0.55em} 0.6184     & \hspace{-0.55em}                      & \hspace{-0.55em}                      & \hspace{-0.55em} \textbf{115} & \hspace{-0.55em} 56.59  & \hspace{-0.55em} \textbf{0.6707} & \hspace{-0.55em} \textbf{}            & \hspace{-0.55em} 78        & \hspace{-0.55em} 46.95   & \hspace{-0.55em} 0.6484     \\
PMS2022                    & \hspace{-0.55em} 179                      & \hspace{-0.55em} \textbf{135} & \hspace{-0.55em} 18.13 & \hspace{-0.55em} \textbf{0.7167} & \hspace{-0.55em} \textbf{}            & \hspace{-0.55em} 72       & \hspace{-0.55em} 11.18   & \hspace{-0.55em} 0.6706     & \hspace{-0.55em}                      & \hspace{-0.55em}                      & \hspace{-0.55em} \textbf{134} & \hspace{-0.55em} 59.33  & \hspace{-0.55em} \textbf{0.7281} & \hspace{-0.55em} \textbf{}            & \hspace{-0.55em} 90        & \hspace{-0.55em} 53.92   & \hspace{-0.55em} 0.7036     \\
PMS2023                    & \hspace{-0.55em} 179                      & \hspace{-0.55em} \textbf{124} & \hspace{-0.55em} 18.26 & \hspace{-0.55em} \textbf{0.6805} & \hspace{-0.55em} \textbf{}            & \hspace{-0.55em} 57       & \hspace{-0.55em} 15.02   & \hspace{-0.55em} 0.6677     & \hspace{-0.55em}                      & \hspace{-0.55em}                      & \hspace{-0.55em} \textbf{128} & \hspace{-0.55em} 85.50  & \hspace{-0.55em} \textbf{0.7130} & \hspace{-0.55em} \textbf{}            & \hspace{-0.55em} 66        & \hspace{-0.55em} 84.85   & \hspace{-0.55em} 0.7087     \\
WPMS2018                   & \hspace{-0.55em} 172                      & \hspace{-0.55em} \textbf{136} & \hspace{-0.55em} 19.32 & \hspace{-0.55em} \textbf{0.7815} & \hspace{-0.55em} \textbf{}            & \hspace{-0.55em} 30       & \hspace{-0.55em} 16.04   & \hspace{-0.55em} 0.7294     & \hspace{-0.55em}                      & \hspace{-0.55em}                      & \hspace{-0.55em} \textbf{139} & \hspace{-0.55em} 90.30  & \hspace{-0.55em} \textbf{0.7939} & \hspace{-0.55em} \textbf{}            & \hspace{-0.55em} 29        & \hspace{-0.55em} 81.13   & \hspace{-0.55em} 0.7611     \\
WPMS2019                   & \hspace{-0.55em} 297                      & \hspace{-0.55em} \textbf{244} & \hspace{-0.55em} 21.02 & \hspace{-0.55em} \textbf{0.7711} & \hspace{-0.55em} \textbf{}            & \hspace{-0.55em} 47       & \hspace{-0.55em} 10.05   & \hspace{-0.55em} 0.6618     & \hspace{-0.55em}                      & \hspace{-0.55em}                      & \hspace{-0.55em} \textbf{252} & \hspace{-0.55em} 94.37  & \hspace{-0.55em} \textbf{0.7983} & \hspace{-0.55em} \textbf{}            & \hspace{-0.55em} 41        & \hspace{-0.55em} 50.19   & \hspace{-0.55em} 0.7060     \\
WPMS2020                   & \hspace{-0.55em} 253                      & \hspace{-0.55em} \textbf{210} & \hspace{-0.55em} 23.94 & \hspace{-0.55em} \textbf{0.7795} & \hspace{-0.55em} \textbf{}            & \hspace{-0.55em} 29       & \hspace{-0.55em} 8.89    & \hspace{-0.55em} 0.6705     & \hspace{-0.55em}                      & \hspace{-0.55em}                      & \hspace{-0.55em} \textbf{221} & \hspace{-0.55em} 106.80 & \hspace{-0.55em} \textbf{0.8125} & \hspace{-0.55em} \textbf{}            & \hspace{-0.55em} 25        & \hspace{-0.55em} 45.01   & \hspace{-0.55em} 0.7162     \\
WPMS2021                   & \hspace{-0.55em} 151                      & \hspace{-0.55em} \textbf{115} & \hspace{-0.55em} 26.53 & \hspace{-0.55em} \textbf{0.6887} & \hspace{-0.55em} \textbf{}            & \hspace{-0.55em} 25       & \hspace{-0.55em} 21.81   & \hspace{-0.55em} 0.5889     & \hspace{-0.55em}                      & \hspace{-0.55em}                      & \hspace{-0.55em} \textbf{122} & \hspace{-0.55em} 127.78 & \hspace{-0.55em} \textbf{0.7252} & \hspace{-0.55em} \textbf{}            & \hspace{-0.55em} 25        & \hspace{-0.55em} 77.79   & \hspace{-0.55em} 0.6410     \\
WPMS2022                   & \hspace{-0.55em} 197                      & \hspace{-0.55em} \textbf{149} & \hspace{-0.55em} 23.59 & \hspace{-0.55em} \textbf{0.7119} & \hspace{-0.55em} \textbf{}            & \hspace{-0.55em} 26       & \hspace{-0.55em} 18.51   & \hspace{-0.55em} 0.6316     & \hspace{-0.55em}                      & \hspace{-0.55em}                      & \hspace{-0.55em} \textbf{163} & \hspace{-0.55em} 125.04 & \hspace{-0.55em} \textbf{0.7613} & \hspace{-0.55em} \textbf{}            & \hspace{-0.55em} 22        & \hspace{-0.55em} 69.88   & \hspace{-0.55em} 0.6790     \\
WPMS2023                   & \hspace{-0.55em} 160                      & \hspace{-0.55em} \textbf{119} & \hspace{-0.55em} 22.73 & \hspace{-0.55em} \textbf{0.6655} & \hspace{-0.55em} \textbf{}            & \hspace{-0.55em} 22       & \hspace{-0.55em} 19.72   & \hspace{-0.55em} 0.5739     & \hspace{-0.55em}                      & \hspace{-0.55em}                      & \hspace{-0.55em} \textbf{126} & \hspace{-0.55em} 110.30 & \hspace{-0.55em} \textbf{0.7125} & \hspace{-0.55em} \textbf{}            & \hspace{-0.55em} 21        & \hspace{-0.55em} 63.33   & \hspace{-0.55em} 0.6361    \\ \bottomrule
\end{tabular}
\caption{Comparison of \name and NuWLS under two time limits.
}
\label{table-NuWLS}
\end{table*}
\begin{table*}[t]
\footnotesize
\centering
\begin{tabular}{lrrrrrrrrr|rrrrrrrr} \toprule
\multirow{2}{*}{Benchmark} &  \multirow{2}{*}{\hspace{-0.55em}\#inst.} &  \multicolumn{3}{c}{\name (60s)}     & \multicolumn{1}{c}{\hspace{-0.55em}} &  \multicolumn{3}{c}{BandMaxSAT (60s)} &  \multicolumn{1}{c}{\hspace{-0.55em}} &  \multicolumn{1}{|c}{\hspace{-0.55em}} &  \multicolumn{3}{c}{\name (300s)}     &  \multicolumn{1}{c}{\hspace{-0.55em}} &  \multicolumn{3}{c}{BandMaxSAT (300s)} \\ \cline{3-5} \cline{7-9} \cline{12-14} \cline{16-18} 
                           & \hspace{-0.55em} \hspace{-0.55em}                          & \hspace{-0.55em} \hspace{-0.55em} \#win       & \hspace{-0.55em} \hspace{-0.55em} time  & \hspace{-0.55em} \hspace{-0.55em} \#score        & \hspace{-0.55em} \hspace{-0.55em}                      & \hspace{-0.55em} \hspace{-0.55em} \#win   & \hspace{-0.55em} \hspace{-0.55em} time    & \hspace{-0.55em} \hspace{-0.55em} \#score   & \hspace{-0.55em} \hspace{-0.55em}                      & \hspace{-0.55em} \hspace{-0.55em}                      & \hspace{-0.55em} \hspace{-0.55em} \#win       & \hspace{-0.55em} \hspace{-0.55em} time   & \hspace{-0.55em} \hspace{-0.55em} \#score        & \hspace{-0.55em} \hspace{-0.55em}                      & \hspace{-0.55em} \hspace{-0.55em} \#win    & \hspace{-0.55em} \hspace{-0.55em} time    & \hspace{-0.55em} \hspace{-0.55em} \#score   \\ \hline
PMS2018                    & \hspace{-0.55em} 153                      & \hspace{-0.55em} \textbf{122} & \hspace{-0.55em} 15.40 & \hspace{-0.55em} \textbf{0.7562} & \hspace{-0.55em} \textbf{}            & \hspace{-0.55em} 49         & \hspace{-0.55em} 15.27     & \hspace{-0.55em} 0.6556      & \hspace{-0.55em}                      & \hspace{-0.55em}                      & \hspace{-0.55em} \textbf{122} & \hspace{-0.55em} 64.57  & \hspace{-0.55em} \textbf{0.7819} & \hspace{-0.55em} \textbf{}            & \hspace{-0.55em} 61         & \hspace{-0.55em} 60.44      & \hspace{-0.55em} 0.6913      \\
PMS2019                    & \hspace{-0.55em} 299                      & \hspace{-0.55em} \textbf{229} & \hspace{-0.55em} 13.49 & \hspace{-0.55em} \textbf{0.7384} & \hspace{-0.55em} \textbf{}            & \hspace{-0.55em} 109        & \hspace{-0.55em} 12.10     & \hspace{-0.55em} 0.6634      & \hspace{-0.55em}                      & \hspace{-0.55em}                      & \hspace{-0.55em} \textbf{238} & \hspace{-0.55em} 57.65  & \hspace{-0.55em} \textbf{0.7701} & \hspace{-0.55em} \textbf{}            & \hspace{-0.55em} 125        & \hspace{-0.55em} 48.99      & \hspace{-0.55em} 0.6973      \\
PMS2020                    & \hspace{-0.55em} 262                      & \hspace{-0.55em} \textbf{196} & \hspace{-0.55em} 14.49 & \hspace{-0.55em} \textbf{0.7378} & \hspace{-0.55em} \textbf{}            & \hspace{-0.55em} 91         & \hspace{-0.55em} 12.74     & \hspace{-0.55em} 0.6744      & \hspace{-0.55em}                      & \hspace{-0.55em}                      & \hspace{-0.55em} \textbf{202} & \hspace{-0.55em} 63.03  & \hspace{-0.55em} \textbf{0.7586} & \hspace{-0.55em} \textbf{}            & \hspace{-0.55em} 104        & \hspace{-0.55em} 49.95      & \hspace{-0.55em} 0.6961      \\
PMS2021                    & \hspace{-0.55em} 155                      & \hspace{-0.55em} \textbf{110} & \hspace{-0.55em} 16.80 & \hspace{-0.55em} \textbf{0.6565} & \hspace{-0.55em} \textbf{}            & \hspace{-0.55em} 57         & \hspace{-0.55em} 11.27     & \hspace{-0.55em} 0.5901      & \hspace{-0.55em}                      & \hspace{-0.55em}                      & \hspace{-0.55em} \textbf{110} & \hspace{-0.55em} 54.66  & \hspace{-0.55em} \textbf{0.6707} & \hspace{-0.55em} \textbf{}            & \hspace{-0.55em} 64         & \hspace{-0.55em} 43.54      & \hspace{-0.55em} 0.6013      \\
PMS2022                    & \hspace{-0.55em} 179                      & \hspace{-0.55em} \textbf{123} & \hspace{-0.55em} 17.95 & \hspace{-0.55em} \textbf{0.7167} & \hspace{-0.55em} \textbf{}            & \hspace{-0.55em} 61         & \hspace{-0.55em} 15.61     & \hspace{-0.55em} 0.6383      & \hspace{-0.55em}                      & \hspace{-0.55em}                      & \hspace{-0.55em} \textbf{123} & \hspace{-0.55em} 57.79  & \hspace{-0.55em} \textbf{0.7281} & \hspace{-0.55em} \textbf{}            & \hspace{-0.55em} 73         & \hspace{-0.55em} 61.39      & \hspace{-0.55em} 0.6574      \\
PMS2023                    & \hspace{-0.55em} 179                      & \hspace{-0.55em} \textbf{133} & \hspace{-0.55em} 18.25 & \hspace{-0.55em} \textbf{0.6805} & \hspace{-0.55em} \textbf{}            & \hspace{-0.55em} 42         & \hspace{-0.55em} 16.28     & \hspace{-0.55em} 0.5768      & \hspace{-0.55em}                      & \hspace{-0.55em}                      & \hspace{-0.55em} \textbf{139} & \hspace{-0.55em} 86.56  & \hspace{-0.55em} \textbf{0.7130} & \hspace{-0.55em} \textbf{}            & \hspace{-0.55em} 46         & \hspace{-0.55em} 75.12      & \hspace{-0.55em} 0.6252      \\
WPMS2018                   & \hspace{-0.55em} 172                      & \hspace{-0.55em} \textbf{126} & \hspace{-0.55em} 18.56 & \hspace{-0.55em} \textbf{0.7815} & \hspace{-0.55em} \textbf{}            & \hspace{-0.55em} 43         & \hspace{-0.55em} 23.14     & \hspace{-0.55em} 0.7123      & \hspace{-0.55em}                      & \hspace{-0.55em}                      & \hspace{-0.55em} \textbf{127} & \hspace{-0.55em} 82.44  & \hspace{-0.55em} \textbf{0.7939} & \hspace{-0.55em} \textbf{}            & \hspace{-0.55em} 41         & \hspace{-0.55em} 79.46      & \hspace{-0.55em} 0.7289      \\
WPMS2019                   & \hspace{-0.55em} 297                      & \hspace{-0.55em} \textbf{228} & \hspace{-0.55em} 20.07 & \hspace{-0.55em} \textbf{0.7711} & \hspace{-0.55em} \textbf{}            & \hspace{-0.55em} 82         & \hspace{-0.55em} 18.79     & \hspace{-0.55em} 0.6751      & \hspace{-0.55em}                      & \hspace{-0.55em}                      & \hspace{-0.55em} \textbf{229} & \hspace{-0.55em} 87.21  & \hspace{-0.55em} \textbf{0.7983} & \hspace{-0.55em} \textbf{}            & \hspace{-0.55em} 94         & \hspace{-0.55em} 71.46      & \hspace{-0.55em} 0.7215      \\
WPMS2020                   & \hspace{-0.55em} 253                      & \hspace{-0.55em} \textbf{196} & \hspace{-0.55em} 24.18 & \hspace{-0.55em} \textbf{0.7795} & \hspace{-0.55em} \textbf{}            & \hspace{-0.55em} 54         & \hspace{-0.55em} 16.38     & \hspace{-0.55em} 0.6382      & \hspace{-0.55em}                      & \hspace{-0.55em}                      & \hspace{-0.55em} \textbf{199} & \hspace{-0.55em} 98.11  & \hspace{-0.55em} \textbf{0.8125} & \hspace{-0.55em} \textbf{}            & \hspace{-0.55em} 61         & \hspace{-0.55em} 63.22      & \hspace{-0.55em} 0.6947      \\
WPMS2021                   & \hspace{-0.55em} 151                      & \hspace{-0.55em} \textbf{99}  & \hspace{-0.55em} 27.37 & \hspace{-0.55em} \textbf{0.6887} & \hspace{-0.55em} \textbf{}            & \hspace{-0.55em} 39         & \hspace{-0.55em} 25.35     & \hspace{-0.55em} 0.5571      & \hspace{-0.55em}                      & \hspace{-0.55em}                      & \hspace{-0.55em} \textbf{92}  & \hspace{-0.55em} 128.10 & \hspace{-0.55em} \textbf{0.7252} & \hspace{-0.55em} \textbf{}            & \hspace{-0.55em} 50         & \hspace{-0.55em} 114.34     & \hspace{-0.55em} 0.6276      \\
WPMS2022                   & \hspace{-0.55em} 197                      & \hspace{-0.55em} \textbf{132} & \hspace{-0.55em} 24.03 & \hspace{-0.55em} \textbf{0.7119} & \hspace{-0.55em} \textbf{}            & \hspace{-0.55em} 48         & \hspace{-0.55em} 25.35     & \hspace{-0.55em} 0.6069      & \hspace{-0.55em}                      & \hspace{-0.55em}                      & \hspace{-0.55em} \textbf{130} & \hspace{-0.55em} 122.92 & \hspace{-0.55em} \textbf{0.7613} & \hspace{-0.55em} \textbf{}            & \hspace{-0.55em} 57         & \hspace{-0.55em} 102.21     & \hspace{-0.55em} 0.6746      \\
WPMS2023                   & \hspace{-0.55em} 160                      & \hspace{-0.55em} \textbf{94}  & \hspace{-0.55em} 23.15 & \hspace{-0.55em} \textbf{0.6655} & \hspace{-0.55em} \textbf{}            & \hspace{-0.55em} 45         & \hspace{-0.55em} 27.42     & \hspace{-0.55em} 0.5795      & \hspace{-0.55em}                      & \hspace{-0.55em}                      & \hspace{-0.55em} \textbf{95}  & \hspace{-0.55em} 107.11 & \hspace{-0.55em} \textbf{0.7125} & \hspace{-0.55em} \textbf{}            & \hspace{-0.55em} 50         & \hspace{-0.55em} 121.94     & \hspace{-0.55em} 0.6379    \\ \bottomrule
\end{tabular}
\caption{Comparison of \name and BandMaxSAT under two time limits. 
}
\label{table-BandMaxSAT}
\end{table*}
\begin{table}[!t]
\centering
\begin{tabular}{lccccc} \toprule
\multirow{2}{*}{Benchmark} &  \multirow{2}{*}{\hspace{-0.55em}\#inst.} &  \multicolumn{2}{c}{\hspace{-0.55em}\name-c}               &  \multicolumn{2}{c}{\hspace{-0.55em}NuWLS-c-2023}  \\ \cline{3-6}
                           & \hspace{-0.55em}                          & \hspace{-0.55em} 60s             & \hspace{-0.55em} 300s                       & \hspace{-0.55em} 60s             & \hspace{-0.55em} 300s            \\ \hline
PMS2018                    & \hspace{-0.55em} 153                      & \hspace{-0.55em} \textbf{0.8477} & \hspace{-0.55em} \textbf{0.8856}            & \hspace{-0.55em} 0.8339          & \hspace{-0.55em} 0.8774          \\
PMS2019                    & \hspace{-0.55em} 299                      & \hspace{-0.55em} \textbf{0.8893} & \hspace{-0.55em} \textbf{0.9349}            & \hspace{-0.55em} 0.8876          & \hspace{-0.55em} 0.9272          \\
PMS2020                    & \hspace{-0.55em} 262                      & \hspace{-0.55em} \textbf{0.8610} & \hspace{-0.55em} \textbf{0.9057}            & \hspace{-0.55em} 0.8606          & \hspace{-0.55em} 0.9007          \\
PMS2021                    & \hspace{-0.55em} 155                      & \hspace{-0.55em} \textbf{0.8403} & \hspace{-0.55em} 0.8947                     & \hspace{-0.55em} 0.8386          & \hspace{-0.55em} \textbf{0.8962} \\
PMS2022                    & \hspace{-0.55em} 179                      & \hspace{-0.55em} \textbf{0.8362} & \hspace{-0.55em} \textbf{0.9042}            & \hspace{-0.55em} 0.8301          & \hspace{-0.55em} 0.8901          \\
PMS2023                    & \hspace{-0.55em} 179                      & \hspace{-0.55em} 0.7933          & \hspace{-0.55em} 0.8754                     & \hspace{-0.55em} \textbf{0.8037} & \hspace{-0.55em} \textbf{0.8780} \\
WPMS2018                   & \hspace{-0.55em} 172                      & \hspace{-0.55em} \textbf{0.8859} & \hspace{-0.55em} \textbf{0.9219}            & \hspace{-0.55em} 0.8774          & \hspace{-0.55em} 0.9210          \\
WPMS2019                   & \hspace{-0.55em} 297                      & \hspace{-0.55em} \textbf{0.8634} & \hspace{-0.55em} 0.9253                     & \hspace{-0.55em} 0.8587          & \hspace{-0.55em} \textbf{0.9259} \\
WPMS2020                   & \hspace{-0.55em} 253                      & \hspace{-0.55em} \textbf{0.8651} & \hspace{-0.55em} \textbf{0.9262}            & \hspace{-0.55em} 0.8582          & \hspace{-0.55em} 0.9233          \\
WPMS2021                   & \hspace{-0.55em} 151                      & \hspace{-0.55em} \textbf{0.7902} & \hspace{-0.55em} 0.8536                     & \hspace{-0.55em} 0.7887          & \hspace{-0.55em} \textbf{0.8569} \\
WPMS2022                   & \hspace{-0.55em} 197                      & \hspace{-0.55em} \textbf{0.7930} & \hspace{-0.55em} \textbf{0.8715}            & \hspace{-0.55em} 0.7851          & \hspace{-0.55em} 0.8711          \\
WPMS2023                   & \hspace{-0.55em} 160                      & \hspace{-0.55em} \textbf{0.8000} & \hspace{-0.55em} \textbf{0.8803} & \hspace{-0.55em} 0.7843          & \hspace{-0.55em} 0.8776 \\ \bottomrule
\end{tabular}
\caption{Comparison of \name-c and NuWLS-c-2023 under two time limits.
}
\label{table-NuWLS-c}
\end{table}
\begin{table*}[t]
\footnotesize
\centering
\begin{tabular}{lrrrrrrrrr|rrrrrrrr} \toprule
\multirow{2}{*}{Benchmark} &  \multirow{2}{*}{\hspace{-0.55em}\#inst.} &  \multicolumn{3}{c}{\name (60s)}     & \multicolumn{1}{c}{\hspace{-0.55em}} &  \multicolumn{3}{c}{\hspace{-1em}SPB-MaxSAT$_{\delta=1}$ (60s)} &  \multicolumn{1}{c}{\hspace{-0.55em}} &  \multicolumn{1}{|c}{\hspace{-0.55em}} &  \multicolumn{3}{c}{\name (300s)}     &  \multicolumn{1}{c}{\hspace{-0.55em}} &  \multicolumn{3}{c}{\hspace{-1em}SPB-MaxSAT$_{\delta=1}$ (300s)} \\ \cline{3-5} \cline{7-9} \cline{12-14} \cline{16-18} 
                           & \hspace{-0.55em} \hspace{-0.55em}                          & \hspace{-0.55em} \hspace{-0.55em} \#win       & \hspace{-0.55em} \hspace{-0.55em} time  & \hspace{-0.55em} \hspace{-0.55em} \#score        & \hspace{-0.55em} \hspace{-0.55em}                      & \hspace{-0.55em} \hspace{-0.55em} \#win   & \hspace{-0.55em} \hspace{-0.55em} time    & \hspace{-0.55em} \hspace{-1em} \#score   & \hspace{-0.55em} \hspace{-0.55em}                      & \hspace{-0.55em} \hspace{-0.55em}                      & \hspace{-0.55em} \hspace{-0.55em} \#win       & \hspace{-0.55em} \hspace{-0.55em} time   & \hspace{-0.55em} \hspace{-0.55em} \#score        & \hspace{-0.55em} \hspace{-0.55em}                      & \hspace{-0.55em} \hspace{-0.55em} \#win    & \hspace{-0.55em} \hspace{-0.55em} time    & \hspace{-0.55em} \hspace{-1em} \#score   \\ \hline
PMS2018                    & \hspace{-0.6em} 153                      & \hspace{-0.6em} \textbf{104} & \hspace{-0.6em} 14.38 & \hspace{-0.6em} \textbf{0.7562} & \hspace{-0.6em} \textbf{}            & \hspace{-0.6em} 96           & \hspace{-0.6em} 12.60 & \hspace{-1em} 0.7432          & \hspace{-0.6em}                      & \hspace{-0.6em}                      & \hspace{-0.6em} \textbf{106} & \hspace{-0.6em} 57.60  & \hspace{-0.6em} \textbf{0.7819} & \hspace{-0.6em} \textbf{}            & \hspace{-0.6em} 103             & \hspace{-0.6em} 54.96    & \hspace{-1em} 0.7708     \\
PMS2019                    & \hspace{-0.6em} 299                      & \hspace{-0.6em} \textbf{206} & \hspace{-0.6em} 12.76 & \hspace{-0.6em} \textbf{0.7384} & \hspace{-0.6em} \textbf{}            & \hspace{-0.6em} 193          & \hspace{-0.6em} 10.05 & \hspace{-1em} 0.7383          & \hspace{-0.6em}                      & \hspace{-0.6em}                      & \hspace{-0.6em} \textbf{213} & \hspace{-0.6em} 51.21  & \hspace{-0.6em} \textbf{0.7701} & \hspace{-0.6em} \textbf{}            & \hspace{-0.6em} 202             & \hspace{-0.6em} 43.33    & \hspace{-1em} 0.7657     \\
PMS2020                    & \hspace{-0.6em} 262                      & \hspace{-0.6em} \textbf{169} & \hspace{-0.6em} 12.45 & \hspace{-0.6em} \textbf{0.7378} & \hspace{-0.6em} \textbf{}            & \hspace{-0.6em} 168          & \hspace{-0.6em} 12.05 & \hspace{-1em} 0.7366          & \hspace{-0.6em}                      & \hspace{-0.6em}                      & \hspace{-0.6em} \textbf{178} & \hspace{-0.6em} 52.85  & \hspace{-0.6em} \textbf{0.7586} & \hspace{-0.6em} \textbf{}            & \hspace{-0.6em} 175             & \hspace{-0.6em} 47.37    & \hspace{-1em} 0.7545     \\
PMS2021                    & \hspace{-0.6em} 155                      & \hspace{-0.6em} 97           & \hspace{-0.6em} 14.27 & \hspace{-0.6em} \textbf{0.6565} & \hspace{-0.6em} \textbf{}            & \hspace{-0.6em} \textbf{99}  & \hspace{-0.6em} 13.97 & \hspace{-1em} 0.6529          & \hspace{-0.6em}                      & \hspace{-0.6em}                      & \hspace{-0.6em} 97           & \hspace{-0.6em} 45.06  & \hspace{-0.6em} \textbf{0.6707} & \hspace{-0.6em} \textbf{}            & \hspace{-0.6em} \textbf{103}    & \hspace{-0.6em} 39.64    & \hspace{-1em} 0.6658     \\
PMS2022                    & \hspace{-0.6em} 179                      & \hspace{-0.6em} \textbf{118} & \hspace{-0.6em} 15.96 & \hspace{-0.6em} \textbf{0.7167} & \hspace{-0.6em} \textbf{}            & \hspace{-0.6em} 108          & \hspace{-0.6em} 14.96 & \hspace{-1em} 0.7113          & \hspace{-0.6em}                      & \hspace{-0.6em}                      & \hspace{-0.6em} 115          & \hspace{-0.6em} 56.96  & \hspace{-0.6em} \textbf{0.7281} & \hspace{-0.6em} \textbf{}            & \hspace{-0.6em} \textbf{118}    & \hspace{-0.6em} 56.02    & \hspace{-1em} 0.7274     \\
PMS2023                    & \hspace{-0.6em} 179                      & \hspace{-0.6em} \textbf{114} & \hspace{-0.6em} 18.90 & \hspace{-0.6em} \textbf{0.6805} & \hspace{-0.6em} \textbf{}            & \hspace{-0.6em} 91           & \hspace{-0.6em} 14.82 & \hspace{-1em} 0.6672          & \hspace{-0.6em}                      & \hspace{-0.6em}                      & \hspace{-0.6em} \textbf{115} & \hspace{-0.6em} 83.81  & \hspace{-0.6em} \textbf{0.7130} & \hspace{-0.6em} \textbf{}            & \hspace{-0.6em} 103             & \hspace{-0.6em} 75.97    & \hspace{-1em} 0.7052     \\
WPMS2018                   & \hspace{-0.6em} 172                      & \hspace{-0.6em} \textbf{112} & \hspace{-0.6em} 17.72 & \hspace{-0.6em} \textbf{0.7815} & \hspace{-0.6em} \textbf{}            & \hspace{-0.6em} 94           & \hspace{-0.6em} 18.38 & \hspace{-1em} 0.7797          & \hspace{-0.6em}                      & \hspace{-0.6em}                      & \hspace{-0.6em} \textbf{114} & \hspace{-0.6em} 79.72  & \hspace{-0.6em} \textbf{0.7939} & \hspace{-0.6em} \textbf{}            & \hspace{-0.6em} 95              & \hspace{-0.6em} 72.90    & \hspace{-1em} 0.7896     \\
WPMS2019                   & \hspace{-0.6em} 297                      & \hspace{-0.6em} \textbf{195} & \hspace{-0.6em} 18.85 & \hspace{-0.6em} \textbf{0.7711} & \hspace{-0.6em} \textbf{}            & \hspace{-0.6em} 163          & \hspace{-0.6em} 19.74 & \hspace{-1em} 0.7659          & \hspace{-0.6em}                      & \hspace{-0.6em}                      & \hspace{-0.6em} \textbf{185} & \hspace{-0.6em} 77.04  & \hspace{-0.6em} \textbf{0.7983} & \hspace{-0.6em} \textbf{}            & \hspace{-0.6em} 179             & \hspace{-0.6em} 80.77    & \hspace{-1em} 0.7913     \\
WPMS2020                   & \hspace{-0.6em} 253                      & \hspace{-0.6em} \textbf{160} & \hspace{-0.6em} 22.07 & \hspace{-0.6em} \textbf{0.7795} & \hspace{-0.6em} \textbf{}            & \hspace{-0.6em} 129          & \hspace{-0.6em} 23.10 & \hspace{-1em} 0.7756          & \hspace{-0.6em}                      & \hspace{-0.6em}                      & \hspace{-0.6em} \textbf{159} & \hspace{-0.6em} 89.33  & \hspace{-0.6em} \textbf{0.8125} & \hspace{-0.6em} \textbf{}            & \hspace{-0.6em} 140             & \hspace{-0.6em} 95.41    & \hspace{-1em} 0.8045     \\
WPMS2021                   & \hspace{-0.6em} 151                      & \hspace{-0.6em} 78           & \hspace{-0.6em} 20.92 & \hspace{-0.6em} 0.6887          & \hspace{-0.6em}                      & \hspace{-0.6em} \textbf{81}  & \hspace{-0.6em} 25.85 & \hspace{-1em} \textbf{0.6897} & \hspace{-0.6em} \textbf{}            & \hspace{-0.6em} \textbf{}            & \hspace{-0.6em} \textbf{85}  & \hspace{-0.6em} 111.24 & \hspace{-0.6em} \textbf{0.7252} & \hspace{-0.6em} \textbf{}            & \hspace{-0.6em} 79              & \hspace{-0.6em} 107.89   & \hspace{-1em} 0.7231     \\
WPMS2022                   & \hspace{-0.6em} 197                      & \hspace{-0.6em} \textbf{103} & \hspace{-0.6em} 20.41 & \hspace{-0.6em} 0.7119          & \hspace{-0.6em}                      & \hspace{-0.6em} 95           & \hspace{-0.6em} 22.01 & \hspace{-1em} \textbf{0.7121} & \hspace{-0.6em} \textbf{}            & \hspace{-0.6em} \textbf{}            & \hspace{-0.6em} \textbf{108} & \hspace{-0.6em} 109.06 & \hspace{-0.6em} \textbf{0.7613} & \hspace{-0.6em} \textbf{}            & \hspace{-0.6em} 107             & \hspace{-0.6em} 107.62   & \hspace{-1em} 0.7541     \\
WPMS2023                   & \hspace{-0.6em} 160                      & \hspace{-0.6em} \textbf{97}  & \hspace{-0.6em} 19.05 & \hspace{-0.6em} \textbf{0.6655} & \hspace{-0.6em} \textbf{}            & \hspace{-0.6em} 81           & \hspace{-0.6em} 21.90 & \hspace{-1em} 0.6615          & \hspace{-0.6em}                      & \hspace{-0.6em}                      & \hspace{-0.6em} \textbf{103} & \hspace{-0.6em} 94.51  & \hspace{-0.6em} \textbf{0.7125} & \hspace{-0.6em} \textbf{}            & \hspace{-0.6em} 78              & \hspace{-0.6em} 88.39    & \hspace{-1em} 0.7079    \\ \bottomrule
\end{tabular}
\caption{Comparison of \name and SPB-MaxSAT$_{\delta=1}$ under two time limits. 
}
\label{table-Ablation-1}
\end{table*}

\begin{table*}[t]
\footnotesize
\centering
\begin{tabular}{lrrrrrrrrr|rrrrrrrr} \toprule
\multirow{2}{*}{Benchmark} &  \multirow{2}{*}{\hspace{-0.55em}\#inst.} &  \multicolumn{3}{c}{\name (60s)}     & \multicolumn{1}{c}{\hspace{-0.55em}} &  \multicolumn{3}{c}{\hspace{-1em}SPB-MaxSAT$_{all}$ (60s)} &  \multicolumn{1}{c}{\hspace{-0.55em}} &  \multicolumn{1}{|c}{\hspace{-0.55em}} &  \multicolumn{3}{c}{\name (300s)}     &  \multicolumn{1}{c}{\hspace{-0.55em}} &  \multicolumn{3}{c}{\hspace{-1em}SPB-MaxSAT$_{all}$ (300s)} \\ \cline{3-5} \cline{7-9} \cline{12-14} \cline{16-18} 
                           & \hspace{-0.55em} \hspace{-0.55em}                          & \hspace{-0.55em} \hspace{-0.55em} \#win       & \hspace{-0.55em} \hspace{-0.55em} time  & \hspace{-0.55em} \hspace{-0.55em} \#score        & \hspace{-0.55em} \hspace{-0.55em}                      & \hspace{-0.55em} \hspace{-0.55em} \#win   & \hspace{-0.55em} \hspace{-0.55em} time    & \hspace{-0.55em} \hspace{-0.55em} \#score   & \hspace{-0.55em} \hspace{-0.55em}                      & \hspace{-0.55em} \hspace{-0.55em}                      & \hspace{-0.55em} \hspace{-0.55em} \#win       & \hspace{-0.55em} \hspace{-0.55em} time   & \hspace{-0.55em} \hspace{-0.55em} \#score        & \hspace{-0.55em} \hspace{-0.55em}                      & \hspace{-0.55em} \hspace{-0.55em} \#win    & \hspace{-0.55em} \hspace{-0.55em} time    & \hspace{-0.55em} \hspace{-0.55em} \#score   \\ \hline
PMS2018                    & \hspace{-0.55em} 153                      & \hspace{-0.55em} \textbf{105} & \hspace{-0.55em} 14.03 & \hspace{-0.55em} \textbf{0.7562} & \hspace{-0.55em} \textbf{}            & \hspace{-0.55em} 85          & \hspace{-0.55em} 16.95     & \hspace{-1em} 0.7333       & \hspace{-0.55em}                      & \hspace{-0.55em}                      & \hspace{-0.55em} \textbf{114} & \hspace{-0.55em} 61.55  & \hspace{-0.55em} \textbf{0.7819} & \hspace{-0.55em} \textbf{}            & \hspace{-0.55em} 85         & \hspace{-0.55em} 60.24      & \hspace{-1em} 0.7660       \\
PMS2019                    & \hspace{-0.55em} 299                      & \hspace{-0.55em} \textbf{207} & \hspace{-0.55em} 12.37 & \hspace{-0.55em} 0.7384 & \hspace{-0.55em} \textbf{}            & \hspace{-0.55em} 175         & \hspace{-0.55em} 13.27     & \hspace{-1em} \textbf{0.7385}       & \hspace{-0.55em}                      & \hspace{-0.55em}                      & \hspace{-0.55em} \textbf{218} & \hspace{-0.55em} 53.34  & \hspace{-0.55em} \textbf{0.7701} & \hspace{-0.55em} \textbf{}            & \hspace{-0.55em} 180        & \hspace{-0.55em} 52.58      & \hspace{-1em} 0.7632       \\
PMS2020                    & \hspace{-0.55em} 262                      & \hspace{-0.55em} \textbf{178} & \hspace{-0.55em} 12.88 & \hspace{-0.55em} \textbf{0.7378} & \hspace{-0.55em} \textbf{}            & \hspace{-0.55em} 139         & \hspace{-0.55em} 13.25     & \hspace{-1em} 0.7321       & \hspace{-0.55em}                      & \hspace{-0.55em}                      & \hspace{-0.55em} \textbf{187} & \hspace{-0.55em} 57.75  & \hspace{-0.55em} \textbf{0.7586} & \hspace{-0.55em} \textbf{}            & \hspace{-0.55em} 145        & \hspace{-0.55em} 47.77      & \hspace{-1em} 0.7543       \\
PMS2021                    & \hspace{-0.55em} 155                      & \hspace{-0.55em} \textbf{109} & \hspace{-0.55em} 14.39 & \hspace{-0.55em} \textbf{0.6565} & \hspace{-0.55em} \textbf{}            & \hspace{-0.55em} 84          & \hspace{-0.55em} 10.17     & \hspace{-1em} 0.6434      & \hspace{-0.55em}                      & \hspace{-0.55em}                      & \hspace{-0.55em} \textbf{112} & \hspace{-0.55em} 55.38  & \hspace{-0.55em} \textbf{0.6707} & \hspace{-0.55em} \textbf{}            & \hspace{-0.55em} 82         & \hspace{-0.55em} 38.36      & \hspace{-1em} 0.6597       \\
PMS2022                    & \hspace{-0.55em} 179                      & \hspace{-0.55em} \textbf{118} & \hspace{-0.55em} 14.96 & \hspace{-0.55em} \textbf{0.7167} & \hspace{-0.55em} \textbf{}            & \hspace{-0.55em} 100         & \hspace{-0.55em} 16.67     & \hspace{-1em} 0.7088       & \hspace{-0.55em}                      & \hspace{-0.55em}                      & \hspace{-0.55em} \textbf{131} & \hspace{-0.55em} 54.94  & \hspace{-0.55em} \textbf{0.7281} & \hspace{-0.55em} \textbf{}            & \hspace{-0.55em} 97         & \hspace{-0.55em} 40.42      & \hspace{-1em} 0.7190       \\
PMS2023                    & \hspace{-0.55em} 179                      & \hspace{-0.55em} \textbf{119} & \hspace{-0.55em} 16.43 & \hspace{-0.55em} 0.6805 & \hspace{-0.55em} \textbf{}            & \hspace{-0.55em} 85          & \hspace{-0.55em} 20.54     & \hspace{-1em} \textbf{0.6822}       & \hspace{-0.55em}                      & \hspace{-0.55em}                      & \hspace{-0.55em} \textbf{127} & \hspace{-0.55em} 84.44  & \hspace{-0.55em} \textbf{0.7130} & \hspace{-0.55em} \textbf{}            & \hspace{-0.55em} 76         & \hspace{-0.55em} 73.71      & \hspace{-1em} 0.7077       \\
WPMS2018                   & \hspace{-0.55em} 172                      & \hspace{-0.55em} \textbf{110} & \hspace{-0.55em} 16.35 & \hspace{-0.55em} \textbf{0.7815} & \hspace{-0.55em} \textbf{}            & \hspace{-0.55em} 93          & \hspace{-0.55em} 16.56     & \hspace{-0.55em} 0.7788       & \hspace{-0.55em}                      & \hspace{-0.55em}                      & \hspace{-0.55em} \textbf{125} & \hspace{-0.55em} 84.36  & \hspace{-0.55em} \textbf{0.7939} & \hspace{-0.55em} \textbf{}            & \hspace{-0.55em} 87         & \hspace{-0.55em} 59.29      & \hspace{-0.55em} 0.7891       \\
WPMS2019                   & \hspace{-0.55em} 297                      & \hspace{-0.55em} \textbf{194} & \hspace{-0.55em} 18.38 & \hspace{-0.55em} \textbf{0.7711} & \hspace{-0.55em} \textbf{}            & \hspace{-0.55em} 158         & \hspace{-0.55em} 18.16     & \hspace{-0.55em} 0.7623       & \hspace{-0.55em}                      & \hspace{-0.55em}                      & \hspace{-0.55em} \textbf{212} & \hspace{-0.55em} 82.05  & \hspace{-0.55em} \textbf{0.7983} & \hspace{-0.55em} \textbf{}            & \hspace{-0.55em} 158        & \hspace{-0.55em} 71.34      & \hspace{-0.55em} 0.7883       \\
WPMS2020                   & \hspace{-0.55em} 253                      & \hspace{-0.55em} \textbf{158} & \hspace{-0.55em} 21.04 & \hspace{-0.55em} \textbf{0.7795} & \hspace{-0.55em} \textbf{}            & \hspace{-0.55em} 131         & \hspace{-0.55em} 21.75     & \hspace{-0.55em} 0.7782       & \hspace{-0.55em}                      & \hspace{-0.55em}                      & \hspace{-0.55em} \textbf{175} & \hspace{-0.55em} 93.48  & \hspace{-0.55em} \textbf{0.8125} & \hspace{-0.55em} \textbf{}            & \hspace{-0.55em} 122        & \hspace{-0.55em} 85.05      & \hspace{-0.55em} 0.8070       \\
WPMS2021                   & \hspace{-0.55em} 151                      & \hspace{-0.55em} \textbf{89}  & \hspace{-0.55em} 22.80 & \hspace{-0.55em} \textbf{0.6887} & \hspace{-0.55em} \textbf{}            & \hspace{-0.55em} 67          & \hspace{-0.55em} 26.41     & \hspace{-0.55em} 0.6855       & \hspace{-0.55em}                      & \hspace{-0.55em}                      & \hspace{-0.55em} \textbf{99}  & \hspace{-0.55em} 115.98 & \hspace{-0.55em} \textbf{0.7252} & \hspace{-0.55em} \textbf{}            & \hspace{-0.55em} 67         & \hspace{-0.55em} 108.04     & \hspace{-0.55em} 0.7172       \\
WPMS2022                   & \hspace{-0.55em} 197                      & \hspace{-0.55em} \textbf{105} & \hspace{-0.55em} 19.52 & \hspace{-0.55em} \textbf{0.7119} & \hspace{-0.55em} \textbf{}            & \hspace{-0.55em} 97          & \hspace{-0.55em} 22.74     & \hspace{-0.55em} 0.7098       & \hspace{-0.55em}                      & \hspace{-0.55em}                      & \hspace{-0.55em} \textbf{120} & \hspace{-0.55em} 109.43 & \hspace{-0.55em} \textbf{0.7613} & \hspace{-0.55em} \textbf{}            & \hspace{-0.55em} 98         & \hspace{-0.55em} 109.02     & \hspace{-0.55em} 0.7530       \\
WPMS2023                   & \hspace{-0.55em} 160                      & \hspace{-0.55em} \textbf{94}  & \hspace{-0.55em} 17.38 & \hspace{-0.55em} \textbf{0.6655} & \hspace{-0.55em} \textbf{}            & \hspace{-0.55em} 79          & \hspace{-0.55em} 18.71     & \hspace{-0.55em} 0.6619       & \hspace{-0.55em}                      & \hspace{-0.55em}                      & \hspace{-0.55em} \textbf{104} & \hspace{-0.55em} 91.82  & \hspace{-0.55em} \textbf{0.7125} & \hspace{-0.55em} \textbf{}            & \hspace{-0.55em} 80         & \hspace{-0.55em} 95.99      & \hspace{-0.55em} 0.7066    \\ \bottomrule
\end{tabular}
\caption{Comparison of \name and SPB-MaxSAT$_{all}$ under two time limits. 
}
\label{table-Ablation-all}
\end{table*}

\subsection{Experimental Setup}
All the algorithms were implemented in C++ and run on a server using an AMD EPYC 7H12 CPU, running Ubuntu 18.04 Linux operation system. We evaluated the algorithms on all the PMS and WPMS instances from the incomplete track of the six recent MSEs\footnote{https://maxsat-evaluations.github.io/}, i.e., MSE2018 to MSE2023. Note that the benchmarks containing all PMS/WPMS instances from the incomplete track of MSE2023 are named PMS\_2023/WPMS\_2023, and so forth. Each instance was processed once by each algorithm with two time limits, 60 and 300 seconds. This is consistent with the settings of the incomplete track of MSEs.

We adopt two kinds of metrics to compare and evaluate the algorithms. The first one is the number of winning instances, represented by `\#win', which indicates the number of instances in which the algorithm yields the best solution among all the algorithms listed in the table. The metric `\#win' has been widely used in comparing local search MaxSAT algorithms~\cite{Dist,CCEHC,SATLike,SATLike3.0,NuWLS,BandMaxSAT}.

The second one is the scoring function used in the incomplete track of MSEs. The score of a solver for an instance is 0 if the solver cannot find feasible solutions, and $(BKC+1)/(cost(A)+1)$ otherwise, where $A$ is its output feasible solution, and $BKC$ is the best-known cost of the instance. The score of a solver for a benchmark is its average score upon all contained instances, represented by `\#score'. The best results appear in bold in the tables.

Parameters in \name mainly include the BMS parameter $k$, the hard clause dynamic weight increment $h_{inc}$, and the increase proportion $\delta$. Note that parameters $k$ and $h_{inc}$ are also used in the baseline algorithms, and our method only introduces one additional parameter. We adopt an automatic configurator called SMAC3~\cite{SMAC} to tune the parameters based on instances in the incomplete track of MSE2017. Note that SMAC3 is also widely used for tuning the baseline solvers, NuWLS and NuWLS-c-2023. The tuning domains of the above parameters are $k \in [50,100]$, $h_{inc} \in [1,30]$, and $\delta \in [1.0005, 1.002]$. The final settings of these parameters in both \name and \name-c are $k = 53$, $h_{inc} = 1$, and $\delta = 1.00072$ for PMS and $k = 97$, $h_{inc} = 28$, and $\delta = 1.001$ for WPMS. Parameters in variant algorithms of \name in the ablation study are also tuned by SMAC3.

\subsection{Comparison with Local Search Baselines}
The comparisons of \name with NuWLS and BandMaxSAT are summarized in Tables~\ref{table-NuWLS} and~\ref{table-BandMaxSAT}, respectively. From the results, one can see that \name exhibits significantly better performance than the two local search baselines in both PMS and WPMS instances, irrespective of the 60s or 300s time limit, according to the `\#win' and `\#score' metrics. Specifically, the number of `\#win' instances of \name is 47-118\% (resp. 68-217\%) more than NuWLS (resp. BandMaxSAT) for PMS benchmarks and 353-641\% (resp. 84-263\%) more than NuWLS (resp. BandMaxSAT) for WPMS benchmarks. The `\#score' of \name is 1-7\% (resp. 9-18\%) higher than NuWLS (resp. BandMaxSAT) for PMS benchmarks and 4-17\% (resp. 9-24\%) higher than NuWLS (resp. BandMaxSAT) for WPMS benchmarks.

The improvement of \name over the baselines in WPMS instances is more obvious because there are many PMS instances in the MSE benchmarks that are simple and can be solved to optimal for these effective local search algorithms. Moreover, \name and NuWLS actually differ only in their clause weighting system and have the common variable selection strategy and searching framework, and BandMaxSAT has a different searching framework that applies the multi-armed bandit model to help select the search directions. The results show that BandMaxSAT has a better complementarity with \name (i.e., more `\#win' instances) than NuWLS in WPMS benchmarks, indicating that our proposed clause weighting system has a relatively comprehensive improvement compared to that in NuWLS.

In summary, our proposed methods that combine our SPB constraint with clause weighting system and adaptive clause weighting strategy help \name significantly outperform the state-of-the-art local search MaxSAT algorithms.

\subsection{Combining with SAT-based Solver}
The comparison results between \name-c and NuWLS-c-2023 are shown in Table~\ref{table-NuWLS-c}. Both them combine their local search components, i.e., \name and an improvement of NuWLS, with a SAT-based solver, TT-Open-WBO-Inc. They first use a SAT solver to generate a feasible assignment as an initial solution, then call the local search algorithm to continue the search with the initial solution, and finally use the cost value of the best solution found by local search as an upper bound for TT-Open-WBO-Inc. Since the MSE only compares the incomplete solvers using the `\#score' metric, we also follow this convention to compare them using this metric.

The results in Table~\ref{table-NuWLS-c} show that \name-c outperforms NuWLS-c-2023 in 11 out of 12 benchmarks under 60 seconds of time limit and 8 out of 12 benchmarks under 300 seconds of time limit, indicating that \name-c is generally more robust and effective than NuWLS-c-2023. The solvers under 300 seconds of time limit exhibit closer performance because the longer running time of the effective TT-Open-WBO-Inc solver narrows the difference in the performance of the local search algorithms.

\subsection{Ablation Study}
To analyze the effectiveness and rationality of our proposed adaptive clause weighting strategy, we compare the \name algorithm with its two variants. The first one simply sets $\delta=1$, denoted as SPB-MaxSAT$_{\delta=1}$, and the second one applies the adaptive clause weighting strategy to both the hard clauses and the SPB constraint, denoted as SPB-MaxSAT$_{all}$. The comparison results of \name with these two variants are summarized in Tables~\ref{table-Ablation-1} and~\ref{table-Ablation-all}, respectively.

The results show that \name outperforms SPB-MaxSAT$_{\delta=1}$, indicating that our adaptive clause weighting strategy is effective and can improve the performance of the clause weighting system. \name also outperforms SPB-MaxSAT$_{all}$, indicating that we only highlight the soft clauses and the objective function is reasonable and can lead to better feasible solutions. Using the adaptive clause weighting strategy for both hard clauses and the SPB constraint may make the algorithm swing between hard and soft clauses and cannot focus on improving the objective function so as to find better solutions.

\section{Conclusion}
Complete algorithm and local search are two research lines of the MaxSAT solvers. Many studies and solvers have tried to apply local search to improve complete solvers, but there have been few attempts in the opposite direction. In this paper, we investigate the utilization of complete solving techniques in improving local search algorithms and focus on the Soft conflict Pseudo Boolean (SPB) constraint widely used in complete solvers to enforce the algorithm to find solutions better than the best solution found so far. We rethink the usage of the SPB constraint and propose integrating SPB into the clause weighting system of local search. We further propose an adaptive clause weighting strategy that allows the dynamic weights to be adjusted proportionally rather than linearly.

Based on the clause weighting techniques, we propose a new local search algorithm, \name. Extensive experiments demonstrate its excellent performance and also show that the proposed adaptive clause weighting strategy is reasonable and effective. In future work, we will further investigate the utilization of complete solving techniques for local search and the adaptive clause weighting strategy.


\bibliographystyle{named}
\bibliography{ijcai24}

\end{document}